\documentclass[sigconf]{acmart}

\copyrightyear{2021}
\acmYear{2021}
\setcopyright{acmcopyright}\acmConference[HT '21]{Proceedings of the 32nd ACM Conference on Hypertext and Social Media}{August 30-September 2, 2021}{Virtual Event, Ireland}
\acmBooktitle{Proceedings of the 32nd ACM Conference on Hypertext and Social Media (HT '21), August 30-September 2, 2021, Virtual Event, Ireland}
\acmPrice{15.00}
\acmDOI{10.1145/3465336.3475118}
\acmISBN{978-1-4503-8551-0/21/08}

\AtBeginDocument{%
  \providecommand\BibTeX{{%
    \normalfont B\kern-0.5em{\scshape i\kern-0.25em b}\kern-0.8em\TeX}}}

\usepackage{fancyhdr}
\usepackage{afterpage}
\usepackage{balance} 
\usepackage{hyperref}
\begin{document}
\fancyhead{}
\title{Debiasing Multilingual Word Embeddings:\\ A Case Study of Three Indian Languages}


\author{Srijan Bansal}
\authornote{*Both authors contributed equally to this research.}
\affiliation{
  \institution{Dept. of ECE,\\ IIT Kharagpur}
  \state{West Bengal}
  \country{India -- 721302}
  \postcode{721302}
}
\email{srijanbansal97@iitkgp.ac.in}
\author{Vishal Garimella}
\authornotemark[1]
\affiliation{
  \institution{Dept. of CSE,\\ IIT Kharagpur}
  \state{West Bengal}
  \country{India -- 721302}
  \postcode{721302}
}
\email{vishal_g@iitkgp.ac.in}
\author{Ayush Suhane}
\affiliation{
  \institution{Dept. of Mathematics,\\ IIT Kharagpur}
  \state{West Bengal}
  \country{India -- 721302}
  \postcode{721302}
}
\email{ayushsuhane99@iitkgp.ac.in}
\author{Animesh Mukherjee}
\affiliation{
  \institution{Dept. of CSE,\\ IIT Kharagpur}
  \state{West Bengal}
  \country{India -- 721302}
}
\email{animeshm@cse.iitkgp.ac.in}

\if{0}
\email{srijanbansal97@iitkgp.ac.in}
\orcid{1234-5678-9012}
\author{Vishal Garimella}
\authornotemark[1]
\email{vishal_g@iitkgp.ac.in}
\orcid{1234-5678-9012}
\author{Ayush Suhane}
\authornotemark[2]
\email{ayushsuhane99@iitkgp.ac.in}
\orcid{1234-5678-9012}
\author{Animesh Mukherjee}
\authornotemark[3]
\email{animeshm@cse.iitkgp.ac.in}
\orcid{1234-5678-9012}
\email{animeshm@iitkgp.ac.in}
\affiliation{
  \institution{Indian Institute of Technology Kharagpur}
  \streetaddress{Hijli}
  \city{Kharagpur}
  \state{West-Bengal}
  \postcode{721302}
}\fi


\newcommand{\red}[1]{\textcolor{red}{#1}}

\newcommand{\blue}[1]{\textcolor{blue}{#1}}

\begin{abstract}
In this paper, we advance the current state-of-the-art method for debiasing monolingual word embeddings so as to generalize well in a multilingual setting. We consider different methods to quantify bias and different debiasing approaches for monolingual as well as multilingual settings. We demonstrate the significance of our bias-mitigation approach on downstream NLP applications. Our proposed methods establish the state-of-the-art performance for debiasing multilingual embeddings for three Indian languages - Hindi, Bengali, and Telugu in addition to English. We believe that our work will open up new opportunities in building unbiased downstream NLP  applications that are inherently dependent on the quality of the word embeddings used. 

\end{abstract}


\if{0}


\fi
\maketitle
\section{Introduction}

Word embeddings are ubiquitous across many downstream NLP applications. Current approaches in NLP rely on large amounts of training data. Such data is easily available for a resource-rich language like English but poses a major challenge for other languages. Multilingual word embeddings are widely used in numerous downstream NLP applications which represent words from multiple languages in a common vector space such that similar meaning words are close to each other. This allows to leverage advancement in English for improving model's performance on low resource languages using transfer learning \cite{Ammar2016MassivelyMW,ahmad-etal-2019-cross,meng-etal-2019-target, chen-etal-2019-multi-source}.

However, there is an unavoidable bottleneck in the aforementioned pipeline. Lately, word embeddings have been found to be `blatantly sexist' thus introducing a bias in the applications built on top of them~\cite{article}. Using Google-News embeddings for the English language the authors showed how gender-neutral profession words (e.g., programmer, homemaker) get aligned to one of the gender directions for various analogy tasks. To mitigate this problem they introduced different flavours of debiasing algorithms that can quite successfully remove the biases in the word embeddings. Later~\citet{Dev2019AttenuatingBI} showed that using simple linear projections can be more effective in attenuating bias in word vectors than complex debiasing algorithms.


\noindent\textbf{An open question}: While multiple efforts have focused on understanding and mitigating the bias in English word embeddings, less attention is given to understand and mitigate the bias in multilingual embeddings. A pertinent question is how well the above debiasing algorithms work for other non-English languages in a multilingual setting. This question is worth investigating for the following reasons.

\begin{itemize}
\item The semantics of gender words may vary from one language to another. ~\citet{comment2} point out \includegraphics[scale=0.5]{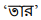} can refer to both `he' or `she' in the sentence \includegraphics[scale=0.5]{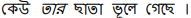} (Somebody forgot his/her umbrella).

\item While~\citet{article} leverages the pronouns (e.g., \textit{she/he}) to construct gendered directions this might not be possible for many languages (e.g., Bengali where the same pronoun is used to refer to both the male and female genders).~\citet{comment} point out that a relatively high proportion of words in Bengali are gender-neutral.

\end{itemize}

\begin{figure*}[h]
    \centering

    \includegraphics[width=\linewidth]{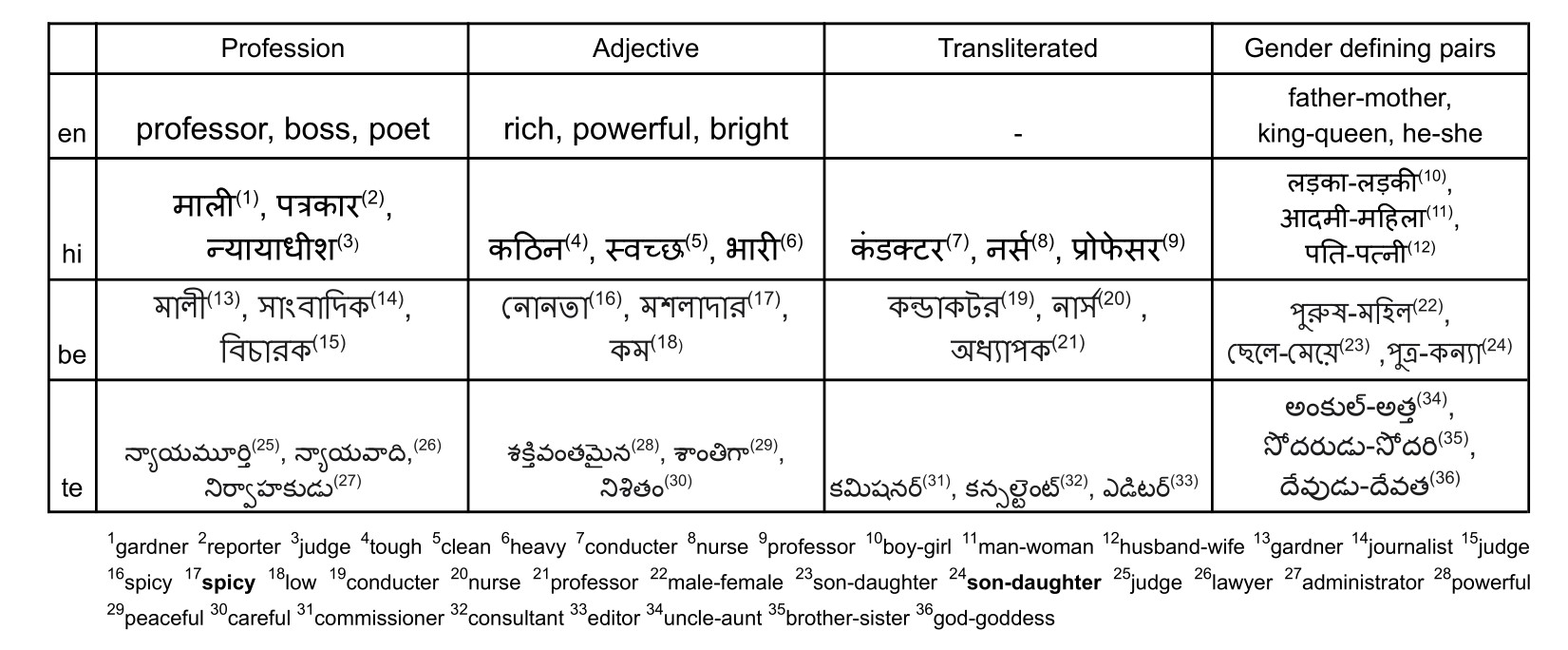}

    \caption{Examples of words in our corpus.}
    \label{gender-words}
\end{figure*}

\noindent\textbf{The present work}: In this paper, we aim to understand bias in multilingual embeddings. Gender bias in English can be very different from bias in other languages due to the innate structure and semantics of the languages. In addition to quantifying this bias, we advance the linear projection based bias mitigation approach to multilingual embeddings.

Our \textbf{key contributions} are as follows.

\begin{itemize}
    \item We build sets of gendered words for three different Indian languages chosen from two different language families (Indo-Aryan and Dravidian) -- Hindi, Bengali and Telugu. We also identify a set of gender-neutral words for these languages which include both profession words (alike~\citet{article}) as well as a set of adjectives which are known to be gender-biased for many Indian languages (see~\citet{anjalipande}).
    \item We perform all our debiasing analysis on fasttext embeddings for the monolingual setting. For multilingual settings, they are aligned to common space using a bilingual dictionary. Our choice of embedding is motivated by the need for mitigating bias in Indic languages as well as in English. 
  \item We advance monolingual debiasing algorithms to multilingual settings and evaluate them based on intrinsic and extrinsic bias measures. Our debiasing approach reduces both intrinsic and extrinsic bias in multilingual setting \textit{achieving state-of-the-art performance} without compromising the overall quality of the debiased embeddings much.
    \end{itemize}

Our methods are generic and can be easily extended to any other language. The choice of the current set of languages is motivated by the knowledge of the authors in these languages. 
All our code is available at \href{https://github.com/amsuhane/Debiasing-Multilingual-Word-Embeddings-A-Case-Study-of-Three-Indian-Languages} \url{https://github.com/amsuhane/Debiasing-Multilingual-Word-Embeddings-A-Case-Study-of-Three-Indian-Languages}.

The rest of the paper is structured as follows. Section~\ref{sec:relwork} describes related works on these problems and provides context on why the problem is difficult and important to solve. Next, in sections~\ref{sec:biasdesc} and~\ref{data}, we respectively quantify bias and describe the datasets which are used to measure it. Sections~\ref{section_4} and~\ref{sec:exp} respectively discuss our novel multilingual debiasing algorithm and the experimental setup. We present the results in section~\ref{sec:results} and conclude our work in section~\ref{sec:conc}.
\section{Related work}\label{sec:relwork}

~\citet{article} provides motivation as to why debiasing word embeddings is a problem of interest and introduce concepts of gender spaces and debiasing with respect to them.~\citet{Dev2019AttenuatingBI} further advance this work by proposing simpler algorithms for debiasing based on linear projections. ~\citet{Caliskan} also find the gender stereotypes in the English word embeddings based on the Word Embedding Association Test (WEAT) which cannot be adapted for other languages. ~\citet{Zhou2019ExaminingGB} reveal that bias exists in languages with grammatical gender and make an attempt to debias English-Spanish embeddings. Recently,~\citet{blodgett-etal-2020-language} have raised concerns about ~\citet{Dev2019AttenuatingBI} and ~\citet{article} due to lack of downstream applications (categorized as ``stereotyping''). In~\cite{Dev_Li_Phillips_Srikumar_2020} the authors studied the advantage of word debiasing on natural language inference task. The authors showed that there is a reduction in the number of invalid inferences due to debiasing. In another recent work~\cite{ravfogel-etal-2020-null} the authors introduce the concept of iterative null space projection to reduce bias in neural representations. The last two works, which are also the most recent in this space, are again limited to the monolingual setting.

Our work is novel in various ways -- we consider multiple Indic languages including samples from both the Indo-Aryan and Dravidian families, perform our analysis after the alignment of words from all the languages to a common space (unlike previous studies) and propose a single algorithm for debiasing all of them which allows for easy inclusion of many other languages. One of the earlier works on debiasing Indo-Aryan languages include~\cite{pujari}. Further, we address the concern of stereotyping by solving a downstream NLP task of occupation classification.

\section{Quantifying Bias}\label{sec:biasdesc}
In this section, we describe the notion of gender bias in word embeddings. Although in this study we consider only four languages namely English, Hindi, Bengali and Telugu the approach is general and can be easily extended to other languages. Bias in embeddings can be quantified in two major ways: \textit{intrinsic} bias and \textit{extrinsic} bias. While the former is used to quantify bias at the word-level the latter focuses on quantifying bias from the perspective of downstream NLP applications. 

\subsection{Intrinsic bias}

Intrinsic bias (\textit{\textit{InBias}}) evaluation metric proposed by ~\citet{zhao2020gender} uses the gap between the distance of occupations and corresponding gender to show gender discrimination. Say, we have a pair of masculine and feminine words describing an occupation, such as the words actor and actress, the only difference lies in the gender information. As a result, they should have similar correlations to the corresponding gender seed words such as `he' and `she'. If there is a gap, i.e., the distance between `actor' and `he' against the distance between `actress' and `she', it means such occupation shows discrimination against gender. We provide detailed descriptions of those words in section \ref{data}.

\noindent{\textbf{Definition of \textit{InBias}}}: Given a set of masculine and feminine words, we define \textit{InBias} as:
\[InBias=\frac{1}{N} \sum_{i=1}^{N}{|dis(O_{M_i},S_M)-dis(O_{F_i},S_F)|}\]
where \[dis(X,Y)=\frac{1}{|Y|} \sum_{y \in Y}{1-cos(X,y)}\]
\normalsize

Here $(O_{M_i}, O_{F_i})$ stand for the masculine and feminine versions of the $i^\textrm{th}$ occupation word, such as (`actor' and `actress'). This metric can also be generalized to languages without grammatical genders, such as English, by just using the same format of the occupation words. In our gender-neutral set ($N_{all}$), we only consider words which are same in male and female versions for all languages. $S_{M}$ and $S_{F}$ denote the set of male and female-oriented words respectively (i.e., $S_{M}$ contains \{male, him, he, $\dots$\}, and $S_{F}$ contains \{female, her, she, $\dots$\}).

In other words, intrinsic bias is a simple geometric method to quantify the overall ``proximity" of gender-neutral words with respect to the notion of gender in a particular embedding space.

\subsection{Extrinsic bias}
The authors in ~\citet{gonen-goldberg-2019-lipstick} raise a valid concern that projection-based debiasing methods such as hard-debias and linear-projection reduce bias superficially. While the bias is indeed substantially reduced according to the provided bias definition, the actual effect is mostly of hiding the bias, not removing it.
 
We address this concern by, quantifying bias from an extrinsic perspective (\textit{ExBias}) based on a downstream task~\cite{zhao2020gender} in addition to the intrinsic metric. 
Extrinsic bias measures the gap in performance for a standard NLP task before and after the debiasing of the word embeddings.

\noindent \textbf{Definition of \textit{ExBias}}: We follow the same definition of extrinsic bias evaluation as used in~\citet{zhao2020gender}, i.e., using the average  performance gap between different gender groups (male and female) aggregated across all the occupations $(|Diff|)$. We used the BiosBias dataset proposed by ~\citet{10.1145/3287560.3287572} to evaluate the bias in predicting the occupation of people using a short biography on the bio of the person written in the third person (see section \ref{data} for more details). We split the dataset based on the gender attribute. A gender-agnostic model should have similar performance in each group. To make predictions of the occupations, we used bidirectional LSTM units followed by attention mechanism (similar to the model used in ~\citet{10.1145/3287560.3287572}). The predictions are generated by a softmax layer. We train such models using standard cross-entropy loss and keep the embeddings frozen during the training.

\section{Dataset}\label{sec:dataset}
\label{data}

In this section we describe the datasets used for measuring the \textit{InBias} and \textit{ExBias}.

\begin{figure*}[t!]
\centering

  \includegraphics[width=\linewidth]{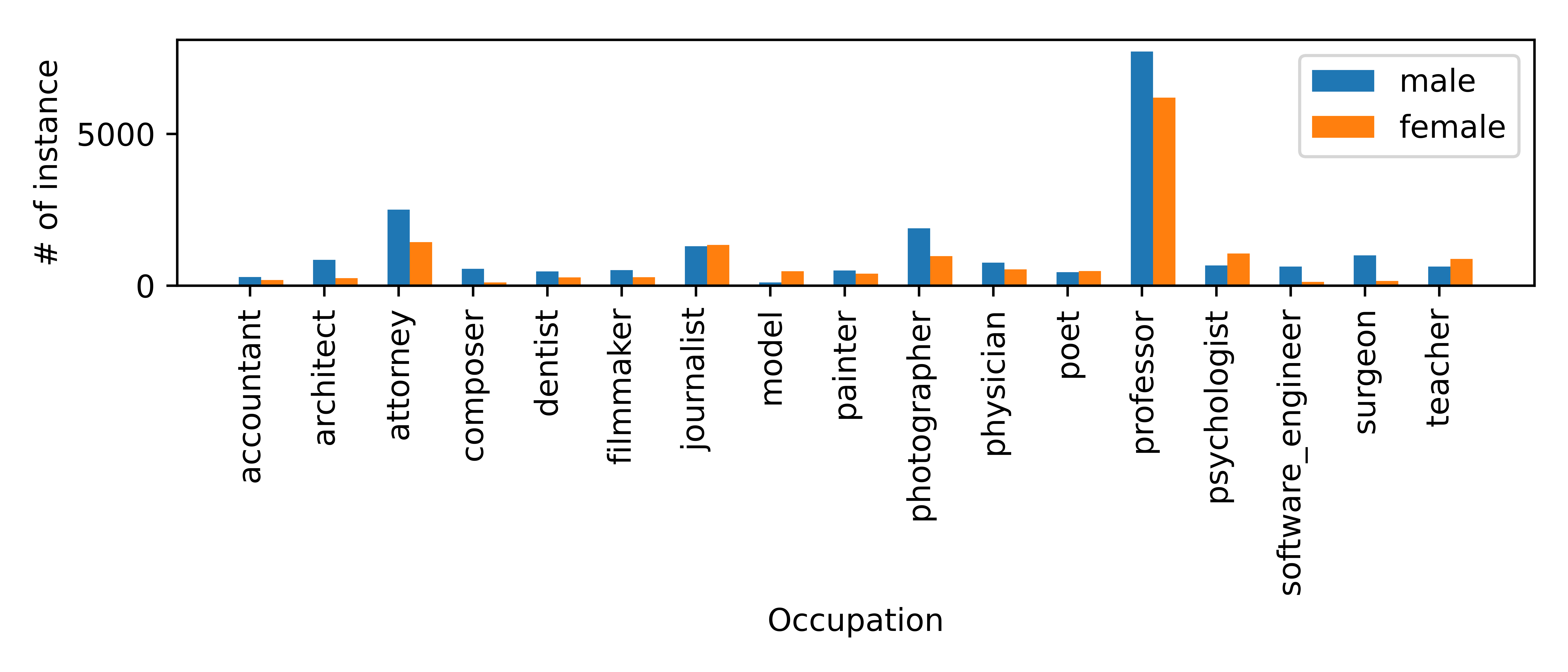}
  \caption{\label{fig4}Gender statistics of Bios-Bias dataset for different occupations where each occupation has at least 100 instances. X-axis here stands for the occupation names and y-axis is the number of instances for each occupation.}
  
\end{figure*}

\subsection{\textit{InBias} dataset} 

In order to evaluate intrinsic bias for embeddings, we require male-female counterparts of words or gender-neutral words and gender-oriented words\footnote{https://bit.ly/2TNKWs3}. We considered a set of gender-neutral occupational words and adjectives ($N_{all}$). For the three Indian languages, it also includes the relevant transliterated-English profession words taken from~\cite{article}. $N_{all}$ consists of neutral words for each language namely $N_{en},N_{hi} ,N_{te},N_{bn}$. 

In addition to neutral words  for each language, we considered a set of gender-defining pairs ($D_{all}$), i.e., gender specific words that are associated with a gender by definition (e.g., \{\{he, she\}, \{him,her\}, \{king,queen\}\}). For $D_{all} = \{D_{all,1},D_{all,2},\dots,D_{all,m}\}$, each $D_{all,i}$ represents a tuple of male-female words (\{he,she\}, \{king,queen\} etc.) $D_{all,i}=(\delta_{i}^{+},\delta_{i}^{-})$ where  $\delta_{i}^{+},\delta_{i}^{-} \in \mathcal{V}$. $D_{all}$ constitutes of $D_{en}$, $D_{hi}$, $D_{be}$ and $D_{te}$ which are gender-defining pairs for each language.  

\begin{table}[h!]
\centering
\begin{tabular}{||c|c|c|c|c|c||} 
    \hline
    Lang & $|N_{prof}|$ & $|N_{adj}|$ & $|N_{tr}|$    & $|N_{lang}|$ & $|D_{lang}|$ \\ [0.4ex] 
    \hline\hline
    $en$ & 59 & 50 & - & 109 & 20\\ 
    $hi$ & 28 & 44 & 14 & 86 & 20\\
    $be$ & 29 & 43 & 15 & 87 & 21\\
    $te$ & 18 & 54 & 18 & 90 & 15 \\
    \hline
    $All$ & 134 & 191 & 47 & \textbf{372} &          \textbf{76}\\[1ex] 
    \hline
    \end{tabular}
    \caption{Datasets statistics: $|N_{prof}|$ - neutral profession words, $|N_{adj}|$ - neutral adjectives, $|N_{tr}|$  - neutral English transliterated words, $|N_{lang}|$ - total number of neutral words for language \textit{lang}, $|D_{lang}|$ - total number of gender pairs for language \textit{lang}.}
    \label{table:1}
    \end{table}
Table~\ref{table:1} notes some basic statistics for the dataset used for further analysis. Our corpus is balanced with respect to different languages. Some examples of words in our dataset are noted in Figure~\ref{gender-words}. Henceforth, `neutral' would refer to gender-neutral words.

\subsection{\textit{ExBias} dataset} 
For the investigation of \textit{\textit{ExBias}} we used the BiosBias dataset proposed by  ~\citet{10.1145/3287560.3287572}. The dataset has been prepared to evaluate the bias in predicting the occupation of people using a short biography on the bio of the person written in third person. We followed the data collection procedure proposed by ~\citet{10.1145/3287560.3287572} for the English dataset\footnote{For our work, we have removed all occurrences of author identity as well as co-references to the authors. We do not plan to release non-anonymized bios-data since it is already available for download from the commoncrawl portal.}. 
 
To identify bio paragraphs, we use regex patterns such as ``NAME is an OCCUPATION-TITLE”. In order to minimise bias due to the dataset, we restricted our analysis to occupations having a relatively balanced distribution of male and female bios as shown in Figure~\ref{fig4}. We extract the binary gender based on gendered pronouns, like `he' and `she'. To remove any effect of names and gender pronouns on the model prediction, we filtered them out from the bios.

One of the main challenges here is the absence of
appropriate datasets (bios) in Indic languages on Common Crawl dumps. As a resort, we translated the English bios to the Indian languages under investigation using the Google Translate API.

\section{Debiasing algorithms}\label{section_4}

~\citet{article} propose a hard-debiasing (HD) method in which the vector difference of corresponding gender-defining pairs (e.g. \{man,woman\}) i.e., $\overrightarrow{man} - \overrightarrow{woman}$ captures the gender direction in the embedding space. They use this gender direction to project the word embeddings of gender-neutral words into a subspace orthogonal to the gender-defining words. The HD method uses a seed set of gender-defining words to train a support vector machine classifier, and use it to expand the initial set of gender-defining words. During the subsequent debiasing process, HD focuses only on gender neutral words (i.e., words not predicted as gender-defining by the classifier). Therefore, if the classifier erroneously predicts a stereotypical word as a gender-defining word, it would not get debiased. 

~\citet{Dev2019AttenuatingBI} proposed linear projection algorithm similar to HD, which also uses a gender-direction to debias. In the rest of the paper, we shall denote the algorithm proposed by~\cite{Dev2019AttenuatingBI} as LP. Both create a vector-space $\mathbf{B} \subset  \mathbb{R}^{d}$ from the top PCA (Principal Component Analysis) component of gender-defining pairs and using this they remove the gender component of the word embeddings. Hard debias removes the component for gender-neutral words only and equalizes the distance of neutral words wrt each gender-defining pair. LP on the other hand achieves debiasing by removing component along $\mathbf{B}$ for all $w \in W$, thus doing away with the need of training an SVM classifier to mine neutral words. This motivates us to use LP as our baseline. 

Formally, for word vector $\vec{w} \in W$\, we denote its projection onto the gender subspace  $\mathbf{B}$ as $\vec{w_B} = \sum_{i=1}^{k}\langle \vec{w},\vec{b_{j}}\rangle \vec{b_{j}} $ where $b_{j}$ are basis for $\mathbf{B}$. Thus, a word vector $\vec{w}$  can be decomposed as $\vec{w} = \vec{w_B} + \vec{w_{\perp}}$, where $\vec{w_{\perp}}$ is the projection onto the orthogonal space~\cite{article}. Note that all the word vectors are normalized to unit normals.

LP removes $\vec{w_B}$ from all the word vectors $\vec{w} \in W$. Thus the component of neutral word along gender subspace is zero thereby, positioning the neutral words to almost equal distance from the gendered words. For a given word vector  $\vec{w} \in \mathbb{R}^d$, the debiased embedding such that $ \vec{{w}'} \in \mathbb{R}^d$ is $\vec{{w}'} = \vec{w} - \vec{w_B}$. If $dim(\mathbf{B})=k$ then resulting debiased embedding
will have dimension $dim(W')= d-dim(\mathbf{B})=d-k$ ($k=1$ if a single direction is used to construct $\mathbf{B}$). 

\subsection{Basis vectors: PCA and PPA}
Let $D$ be any set of gender-defining pairs such that $D \subseteq D_{all}$. We can represent $D$ as  $D = \{D_1,D_2,\dots,D_n\}$. For each $D_i = \{\delta_{i}^{+},\delta_{i}^{-}$\}, \textit{difference vector} can be defined as $\vec{\delta_{i}} = \vec{\delta_{i}^{+}}-\vec{\delta_{i}^{-}}$. We can stack these difference vectors to form a matrix $Q =$ [$\vec{\delta_{1}} \vec{\delta_{2}} \dots \vec{\delta_{n}}$]$^T$. Now, the gender subspace $\mathbf{B}$ can be obtained from $Q$ using the span of the top-$k$ directions obtained from either principal component analysis (PCA) or principal polynomial analysis (PPA -- Projection Pursuit Analysis) (see supplementary information). 
That is
$\mathbf{B} = span \{\vec{b_1},\vec{b_2} ,\dots,\vec{b_k}\}$ where $\vec{b_i} \in \mathbb{R}^d$ for integer parameter $k>1$. We denote $\mathbf{B}_{lang}$ as the space constructed using only the set $D_{lang}$.

\subsection{LP for multilingual embeddings}

Using linear projection for debiasing multilingual word embeddings naturally raises some questions: LP takes the defining set $D$ as an input to construct the gender-subspace $\mathbf{B}$ and gender neutral set $N$ is needed by the evaluation metric to estimate the effect of debiasing. When there are many languages in a common space ($N \subset N_{all}$), we can have different choices of $D$ ($D_{en}$, $D_{hi} \cup D_{en}$ etc.) for a given $N$.
\begin{itemize}
\item Are the gender subspaces induced by gendered pairs for different languages $l_1$ and $l_2$ (say $\mathbf{B}_{l_1}$ and $\mathbf{B}_{l_2}$) semantically different?

\item What is the best choice for the gender-subspace $\mathbf{B}$ among $\mathbf{B}_{l_1}$ and $\mathbf{B}_{l_2}$ for debiasing both $l_{1}$ and $l_{2}$ words in common a space (which may be either aligned to $l_1$ or $l_2$ in the first place).

\item Could there be a better alternative for constructing $\mathbf{B}$ as opposed to using either $\mathbf{B}_{l_1}$ and $\mathbf{B}_{l_2}$? Will it be sufficient to take gender-defining pairs from all language (here both $l_1$ and $l_2$) to form gender-subspace $\mathbf{B}$?

\item What if the gender subspace $\mathbf{B}$ so formed has an over-representation of vectors from one language over others.
\end{itemize}

We proceed to answer these questions in the following. Let $\textbf{B}_{lang} =span\{\vec{b}_{lang,1},\vec{b}_{lang,2},\dots,\vec{b}_{lang,k}\}$ where $\vec{b}_{lang,i}$ is the $i^\textrm{th}$ most significant basis vector of the gender subspace $\mathbf{B}_{lang}$ (PCA/PPA  components of matrix $Q_{lang}$ obtained from $D_{lang}$). We hypothesize that for two languages $l_{1},l_{2} \in \{en,hi,be,te\}$ s.t. $l_{1} \neq l_{2}$, their corresponding gender subspaces $\mathbf{B}_{l_{1}}$ and $\mathbf{B}_{l_{2}}$ are inherently different from each other. These differences may arise due to different gender semantics of each language. 


To further motivate our hypothesis, we use linear projection algorithm for debiasing neutral words $N_{l_{2}} \subset N_{all}$ using gender direction $\vec{b}_{l_{1},1}$ ($\mathbf{B}_{l_{1}}, k=1$) instead of $\vec{b}_{l_{2},1}$ ($\mathbf{B}_{l_{2}}, k=1$). If $\mathbf{B}_{l_{1}}$ and $\mathbf{B}_{l_{2}}$ are semantically similar, then the choice of the gender direction as either $\vec{b}_{l_{1},1}$ or $\vec{b}_{l_{2},1}$ for debiasing $N_{l_{2}}$ does not matter much since debiasing on one ensures minimal to no component along another.

We score the performance using the metric $S_{l_{1},l_{2}}$ defined as: $S_{l_{1},l_{2}} = \frac{1}{|N_{l_{2}}|}\sum_{\substack{i=1 \\ \forall w_{i} \in N_{l_{2}}}}^{|N_{l_{2}}|} \frac{||\langle \vec{w'_{i}},\vec{b}_{l_{2},1}\rangle|-|\langle \vec{w_{i}},\vec{b}_{l_{2},1}\rangle||}{|\langle \vec{w_{i}},\vec{b}_{l_{2},1}\rangle|}$, where $\vec{w'}$ corresponds to the debiased embedding of word $w \in N_{l_2}$ wrt $\vec{b}_{l_{1},1}$  having original word embedding $\vec{w} \in \mathbb{R}^d$. 
This metric captures aggregate relative change in, neutral-word similarity of a language wrt gender direction in that same language before and after debiasing. Note that $\langle \vec{w'_{i}},\vec{b}_{l_{1},1}\rangle = 0$ since embeddings are debiased wrt $\vec{b}_{l_{1},1}$. Now, if $\mathbf{B}_{l_{1}}$ and $\mathbf{B}_{l_{2}}$ are semantically similar $\langle \vec{w'_{i}},\vec{b}_{l_{2},1}\rangle \approx 0$ thus $S_{l_{1},l_{2}} \to 1$ (ideal case: $S_{l_{1},l_{1}} = 1$).

Table~\ref{table:2} shows the cross-language debiasing performance. The results in Table~\ref{table:2} clearly supports our hypothesis that gender subspace $\mathbf{B}_{en}$, $\mathbf{B}_{hi}$, $\mathbf{B}_{te}$ and $\mathbf{B}_{be}$ are significantly different (since the off-diagonal values are far from unity) and the objective of debiasing the multilingual embedding cannot be accomplished using a single language's gender subspace. For example, $\mathbf{B}_{en}$ is \textbf{not} a good choice for $hi$ words even while the embeddings are in the common space. 

\begin{table}[h!]
\centering
\begin{tabular}{||c|c|c|c|c||}
 \hline
 Lang & $N_{en}$ & $N_{hi}$ & $N_{be}$ & $N_{te}$ \\ [0.4ex]
 \hline\hline
 $b_{en}$ & 1.0 & 0.143 & 0.137 & 0.038\\ 
 $b_{hi}$ & 0.105 & 1.0 & 0.083 & 0.023 \\
 $b_{be}$ & 0.345 & 0.126 & 1.0 & 0.075 \\
 $b_{te}$ & 0.049 & 0.054 & 0.157 & 1.0\\
  \hline
\end{tabular}
\caption{$S_{l_{1},l_{2}}$ for debiasing $N_{l_{2}}$ neutral words using gender direction $\vec{b}_{l_{1},1}$ where $\{l_{1},l_{2}\} \in \{en,hi,be,te\}$.}
\label{table:2}

\end{table}

\noindent\textbf{Choice of $k$}: In the traditional setting, LP uses top-1 gender direction for experiments limited to English. However, in a multi-lingual setting $k=1$ would be a bad choice as we have just observed (the gender direction of one language may be ineffective/insufficient in capturing the gender direction of another language). Hence $k>1$ is a natural choice.

Hence for extending LP to a multilingual setting, the gender space $\mathbf{B}_{all}$ should be constructed using gender-defining training pairs drawn from all languages $D_{all}$. PCA/PPA can then used to obtain top-$k$ basis-vectors $\vec{b}_{all,1},\vec{b}_{all,2},\dots,\vec{b}_{all,k}$ of $\mathbf{B}_{all}$ such that $\mathbf{B}_{all} = span\{\vec{b}_{all,1},\vec{b}_{all,2},\dots,\vec{b}_{all,k'}\} $.

In addition, we also use another approach which constructs bias subspace $\mathbf{B}_{equal\_rep}$ ensuring equal representation of each language in the basis vectors of $\mathbf{B}_{equal\_rep}$. For $L$ languages under consideration and gender subspace having $k$ basis vectors s.t. $k\leq |D_{all}|$, we label the PCA/PPA components of $\mathbf{B}_{all}$ according to their language orientation (see supplementary material for details) and choose top $\frac{k}{L}$ components for each language.  
Motivation for this choice stems from the fact that if all the basis vectors are oriented along a single $lang$ in the derived $\mathbf{B}_{all}$, then the gender space spanned by this subspace will behave similar to $\mathbf{B}_{lang}$, defeating the purpose of multilingual gender representation.

Taking top-$k$ PCA components of the gender directions will always provide a more robust estimate of gender than using a single gender direction. We investigate that by changing the size of seed-gender directions $D$ used to construct $Q$. The variance in InBias metrics is considerably small. We also investigate the same when we randomly sample any subset of $D$ (say he-she/ man-woman) and use it as gender direction, the bias-scores indicate a very low variance.
In the rest of the paper, using only $\mathbf{B}_{lang}$ for debiasing words of a single language $lang$ is referred to as $LP_{mono}$. Similarly, using $\mathbf{B}_{all}$ or $\mathbf{B}_{equal\_rep}$ for debiasing is referred to as $LP_{multi}$ or $LP_{EQR}$ respectively.

\if{0}\subsubsection{Does $B_{all}$ represent all languages equally?}

Now given that we take $k$ PCA components to create gender-subspace, we try to answer if it necessarily needs to be top-$k$ components, as different vector spaces can be constructed by taking different basis vectors in multilingual setting(not necessarily top-k).

To check whether ${B}_{all}$ is really representative of all languages,
we first define the language orientation of a direction $\vec{v}$, as language $l$  which has maximum value of $<\vec{v}, \vec{\bar{l}}>$ or $l=argmax <\vec{v}, \vec{\bar{l}}> $.
Here $\vec{\bar{l}}$ is the mean of all gender directions for that language.
we label the directions $\vec{b}_{all,i} \in \mathbf{B}_{all}$ with their language orientation $l_i$  based on their similarity with mean-language-gender-direction $\bar{l}$ for each $l$.

We see that for top PCA components from $D_{all}$ Figure~\ref{fig:lo} different languages contribute in different proportions to the bias subspace which might lead to preference of certain language over others in terms of capturing the gender semantics which can eventually affect the performance of the debiasing algorithm.

\begin{figure*}[h!]
\begin{center}
  \includegraphics[scale=0.43]{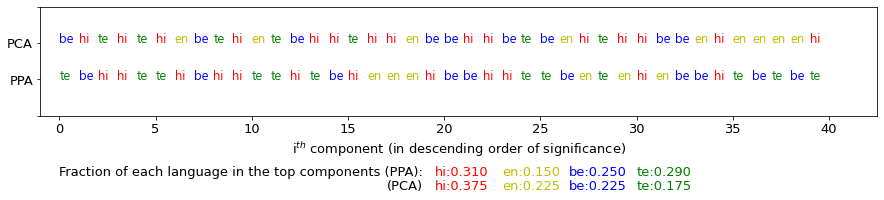}
\end{center}
  \caption{\label{fig:lo}Language orientation $l_i$ of top PCA and PPA components.}
\end{figure*}

 Based on our observations, that gender space of each language is different for each language, we present another variant of LP-multi which constructs bias subspace $\mathbf{B}_{equal\_rep}$  ensuring equal representation of each language in the basis vectors of $B_{equal\_rep}$. For $L$ languages under consideration and gender subspace having $k$ basis vectors s.t. $k\leq |\mathbf{D}_{all}|$, we choose $\frac{k}{L}$ top basis vectors  of having each language-orientation $l_i=l$ for each language (indicated by their orientation labels: $l_{i}$ for $1\leq i \leq k$).
 
 Motivation of this choice stems from the fact that if all basis vectors are language-oriented along $lang$, then the gender space spanned by them will behave similar to $B_{lang}$, defeating the purpose of multilingual gender representation.
 
 From Figure 2 as we see the top PCA-7  components $\vec{b_{all,1}}$,$\vec{b_{all,2}}$,$\vec{b_{all,3}}$,$\vec{b_{all,4}}$,$\vec{b_{all,5}}$,$\vec{b_{all,6}}$,$\vec{b_{all,7}}$ have inclinations be, hi, te ,hi, en, be, te.If we want to construct a 4 dimensional-gender space $k=dim(gender\_space)=4$ then  $B_{all}=span\{b_{all,1},b_{all,2},b_{all,3},b_{all,4} \}$ and $B_{equal\_rep}=span\{b_{all,1},b_{all,2},b_{all,3},b_{all,5} \}$. 
 
 This is essentially a tradeoff in taking top components( which ensure that maximum total gender information is captured) vs performance for under-represented languages in top components.This holds as we have seen that, gender-spaces induced by each language's gender pairs are not semantically similar enough.
 
 This approach restricts $k$ to integral multiples of $L$. We trade-off for equal representation of languages by including less significant components (based on variance or kurtosis) in $\mathbf{B}_{equal\_rep}$ gender subspace.

\fi
\section{Experimental setup}\label{sec:exp}

In our experiments to gender neutralize multilingual word embeddings, we attempt to (i) try kurtosis based projection analysis (PPA) instead of variance based (PCA) and compare them for obtaining top-$k$ components of the gender subspace (ii) compare performances of variants of linear projection algorithm for monolingual and multilingual embeddings for both intrinsic (\textit{InBias}) and extrinsic (\textit{ExBias}) bias metrics.

\subsection{Train test split of gender-defining pairs}
For construction of the gender subspace used by LP we use 10 gender pairs (\textit{train gender pairs}) of each language, i.e., $D_{lang}^{train} \subset D_{lang}$ such that $|D_{lang}^{ train}|=10$ for $lang \in \{en,hi,be,te\}$. Remaining gender pairs (\textit{test gender pairs}) are used as set $D_{lang}^{test} \subset D_{lang}$ such that $D_{lang}^{train} \cap D_{lang}^{test} = \phi$ for \textit{InBias} computation. Note that this split is needed since our evaluation metric is based on cosine similarities with respect to gender directions, and evaluating the embeddings on the same directions as used for constructing gender spaces would typically lead to better values of metrics (train-test leakage). We resort to train-test split strategy for gender pairs because it is intuitive that if the gender subspace obtained using the training gender pairs captures the linguistic concept of gender well, we can expect to get reliable estimate of performance with test gender pairs.

\subsection{Monolingual debiasing}
Monolingual setup characterizes single language applications, where the focus is on debiasing a specific language.  A straightforward way is to use linear projection (LP) to debias embedding using language-specific gender subspace $\mathbf{B}_{lang}$ (say, we expect $hi$ word embeddings to be best debiased when $\mathbf{B}_{hi}$ is used). Thus, for monolingual settings we construct language gender subspace $\mathbf{B}_{lang}$ using gender pairs $D_{lang}$ for a particular language $lang \in \{en,hi,be,te\}$, $k$ basis vectors of $\mathbf{B}_{lang}$ thus obtained using PCA/PPA $(k=4)$ are used to debias monolingual embedding and its performance is evaluated based on \textit{InBias} and \textit{ExBias}.

\subsection {Multilingual debiasing}

Multilingual setup characterizes two or more language applications, where the focus is on debiasing multiple languages at once. We get multilingual embeddings by aligning monolingual embedding in a common space. This is done using bilingual dictionaries between all source-target pairs of languages by~\citet{smith2017offline}'s alignment algorithm (also used in fasttext multilingual pipeline) with same hyper-parameter settings\footnote{\url{https://github.com/babylonhealth/fastText\_multilingual}}. We obtain all these dictionaries from MUSE\footnote{\url{https://github.com/facebookresearch/MUSE\#ground-truth-bilingual-dictionaries}}. Since for Telugu this is not available, we construct Telugu bilingual dictionary wrt other languages using the Google Translate API.

We use the multilingual extension of LP approach, i.e., $LP_{multi}$ and $LP_{EQR}$ (both $k=4$) for debaising. Multilingual gender space $\mathbf{B}_{all}$ is constructed using training gender pairs drawn from all languages $D_{all}^{train}$. PCA/PPA are used to obtain top-$k$ basis vectors of $\mathbf{B}_{all}$ and its performance is evaluated using test gender pairs  $D_{all}^{test}$. Specifically, for bilingual embedding denoted by $l_{1} \rightarrow l_{2}$ (i.e., $l_{1}$ is aligned in $l_{2}$ space), we use both gender-defining pair sets $D_{l_{1}}$ and $D_{l_{2}}$ to construct vector space capturing gender in $l_1 \rightarrow l_2$. This is in turn used by debiasing algorithms to remove the bias component for all words ($l_{1}$ and $l_{2}$) in the vocabulary. We experiment with PCA and PPA methods to get basis vectors for $LP_{multi}$ and $LP_{eqr}$.

We also experimented language-specific (monolingual) LP approaches (denoted as $LP_{l_{1}}$ for language $l_1$) on bilingual embedding $l_{1} \rightarrow l_{2}$ to show multilingual extension of LP is always better than original LP approach. We compared the performance of original and debiased embeddings using metrics \textit{InBias} and \textit{ExBias}.

\subsection{Additional evaluation based on NLP task}
Further, as a quality evaluation of the resulting debiased embeddings we choose an NLP task as well. The usefulness/purpose of word embeddings lies in its utility for a downstream task. To evaluate the quality of word embeddings we considered two downstream tasks. 

For monolingual setting, we test the original and resulting embeddings on a news classification task for all the four languages\footnote{AG news classification (en), BBC news article classification (hi), Soham news article classification (be), iNLTK headline classification (te).}. We used a bidirectional LSTM encoder followed by a softmax layer. Performance is reported in terms of model accuracy.

For multilingual setting, we used the CVIT-\textit{Mann ki baat} dataset \cite{siripragada-etal-2020-multilingual} for the cross-lingual sentence retrieval task. This corpus contains parallel English sentences with Indic languages, thus limiting us to English language pairs like ($hi \rightarrow en$, $en \rightarrow be$ etc). We trained a sentence encoder (bidirectional LSTM) to cross-lingual sentence embedding and used cosine distance for retrieval as suggested in ~\citet{Libovick2019HowLI}. Performance is reported in terms of precision@10. This confirms the possibility to use such embeddings in downstream tasks.
\section{Results}\label{sec:results} 
\subsection{\textit{InBias} results}

We compare intrinsic bias measure for both original fasttext embedding (Orig) and \textit{LP-mono} debiased embeddings (using both PCA and PPA; k=4). Table~\ref{mono} summarizes \textit{InBias} results for monolingual setting (focus only on words in $N_{lang}$), where the training gender subspace is constructed by  $D^{train}_{lang}$ and performance is evaluated using $D^{test}_{lang}$. We observe that $LP_{mono}$ (PCA) outperforms (lower is better) both $LP_{mono}$ (PPA) and original embeddings for all the four languages considered.

\begin{table}[h!]
\centering
\begin{tabular}{|c|c|c|c|}
 \hline
 	$lang$ & Orig & $LP_{mono}^{PCA}$ & $LP_{mono}^{PPA}$\\ [0.4ex]
 \hline\hline
 $en$  & 0.014 & \textbf{0.0} & 0.001 \\   \hline
 $hi$ & 0.015 & \textbf{0.008} & 0.019\\     \hline
 $be$  &0.016 & \textbf{0.011} & 0.023\\    \hline
$te$ & 0.040 & \textbf{0.018} & 0.029 \\   \hline
\end{tabular}
  \caption{\label{mono}\textit{InBias} measure for monolingual setting.}
\end{table}

Table~\ref{multi} summarizes the \textit{InBias} results for the multilingual setting ($N_{all}$), i.e., the gender subspace is constructed using $D^{train}_{all}$ and performance is evaluated using $D^{test}_{all}$. We compare the multilingual extension of LP ($LP_{multi}$ and $LP_{EQR}$) with the $LP_{mono}$ approach. $LP_{mono}$ approach can be applied on a bilingual embedding $l_{1} \rightarrow l_{2}$ by either debiasing it only using language $l_{1}$ or $l_{2}$ (denoted as $LP_{l_{1}}$ and $LP_{l_{2}}$ respectively in Table~\ref{multi}). The interesting observations that one can make from these results are – 
\textbf{(i)} $LP_{multi}$ and $LP_{EQR}$ always perform better than $LP_{mono}$ based approaches ($LP_{l_{1}}$ and $LP_{l_{2}}$) and the original embedding, thus supporting our intuition for multilingual extension of LP,
\textbf{(ii)} $LP_{EQR}$ is at par or slightly better than $LP_{multi}$ for all the bilingual embeddings (except $hi\rightarrow be$), and \textbf{(iii)} PCA based methods always outperform PPA based methods and hence we only show the PCA based in results Table~\ref{multi} to save space (see supplementary material for the PPA based results).

\begin{table}[h!]
\centering
\begin{tabular}{|c|c|c|c|c|c|}
 \hline
 	$l_{1} \rightarrow l_{2}$ & Orig & $LP_{l_1}$ & $LP_{l_2}$ & $LP_{multi}$ & $LP_{EQR}$ \\ [0.4ex]
 \hline\hline
 $en \rightarrow hi$ & 0.013 & 0.015 & 0.010 & \textbf{0.007} & \textbf{0.007}\\
    \hline
 $en \rightarrow be$ & 0.013 & 0.015 & 0.012 & \textbf{0.008} & \textbf{0.008}\\
    \hline
 $en \rightarrow te$ & 0.015 & 0.011 & 0.014 & \textbf{0.006} & \textbf{0.006} \\
    \hline
    
 $hi \rightarrow en$ & 0.015 & 0.013 & 0.013 & \textbf{0.007} & 0.008\\
    \hline
 $hi \rightarrow be$ & 0.016 & 0.015 & 0.013 & 0.008 & \textbf{0.007}\\
    \hline
 $hi \rightarrow te$ & 0.017 & 0.015 & 0.016 & \textbf{0.007} & \textbf{0.007}\\
    \hline
    
 $be \rightarrow en$ & 0.017 & 0.015 & 0.012 & \textbf{0.008} & \textbf{0.008}\\
    \hline
 $be \rightarrow hi$ & 0.016 & 0.014 & 0.014 & \textbf{0.008} & 0.009 \\
    \hline
 $be \rightarrow te$ & 0.015 & 0.014 & 0.013 & \textbf{0.008} & \textbf{0.008} \\
    \hline
    
$te \rightarrow en$ & 0.040 & 0.042  & 0.017 & \textbf{0.013} & 0.014 \\
    \hline
 $te \rightarrow hi$ & 0.024 & 0.019 & 0.018 & \textbf{0.007} & 0.009 \\
    \hline
 $te \rightarrow be$ & 0.019 &  0.023 & 0.022 & \textbf{0.007} & \textbf{0.007} \\
   \hline
  
 \end{tabular}
  \caption{\textit{InBias} measure for multilinugal setting. Debiased algorithm uses PCA with value of $k=4$. 
  }
  \label{multi}
\end{table}

\subsection{\textit{ExBias} results}\label{sec:expbias}

 As discussed earlier for measuring \textit{ExBias} we make use of the BiosBias dataset, whereby, we classify the occupation from a gender-details-scrubbed-bio of a person. The premise for choosing this task is
that the performance (accuracy) of any classification model on the BiosBias dataset must be similar across the male and female applicants, i.e., the accuracy of the classification model on the male applicant bios (denoted by \textbf{M}) must be equivalent to that of female applicant (denoted by \textbf{F}) as the bios are indistinguishable in all other ways. But if occupations themselves have inherent gender connotations, then an asymmetry in the performance is expected. Recall that the absolute difference of accuracy between male and female bios, averaged across all occupations quantifies \textit{ExBias} (i.e., $|Diff|$). The lesser the value of $|Diff|$ the better it is. Also, for a more occupation-wise analysis, we use \textbf{$f_{i}$} to denote the fraction of total occupations for which the performance gap ($|Diff|$) decreases after debiasing.

\begin{table}[h!]
\centering
\begin{tabular}{||c|c|c|c|c|c|c|c||}
 \hline
 $l$ & Emb & Male & Female & $|Diff|$ & $f_i$ & $Acc_{task}$ \\ [0.4ex]
 \hline\hline
 $en$ & orig  & 66.92 & 70.12 & 7.47 & &89.14 \\ 
  & $LP_{mono}$  & 69.69 & 73.63 & \textbf{7.45} & 0.411 & 89.60\\ 
  \hline
 $hi$ & orig & 60.95 & 62.78 & \textbf{6.76} & &45.58\\ 
  & $LP_{mono}$ & 63.13 & 65.76 & 7.76 & 0.647 &41.01\\ 
    \hline
 $be$ & orig & 59.31 & 60.68 & 7.56 & & 84.86\\ 
  & $LP_{mono}$ & 63.53 & 65.58 & \textbf{7.52} & 0.764 & 83.70\\ 
    \hline
 $te$ & orig & 57.97 & 59.39 & 7.50 & & 53.29\\ 
  & $LP_{mono}$  & 63.31 & 64.76 & \textbf{7.28} & 0.353 & 52.11\\ 
   \hline
\end{tabular}
\caption{\label{ex-momo}\textit{ExBias} results in the monolingual settings. 
}
  \label{table:7}  
\end{table}

We compare extrinsic bias (\textit{ExBias}) measure for the original embeddings (orig) and $LP_{mono}$ debiased embeddings\footnote{We always report results for the PCA approach since they outperform the PPA approach.}. Table~\ref{table:7} summarizes the \textit{ExBias} results for the monolingual setting. 
We observe that \textbf{(i)} $LP_{mono}$ always reduces extrinsic bias $(|Diff|)$ without affecting the embedding quality (i.e., the accuracy ($Acc_{task}$) of the news category classification task) much, with the only exception of Hindi, \textit{(ii)} For all the languages, more than 35\% of total professions saw a reduction in performance gap ($f_{i}$) rising up to almost 76\% for Bengali after debiasing, \textbf{(iii)} We also observe that for all the languages, individual male and female accuracy (\textbf{M} and \textbf{F}) increase after debiasing.

\begin{table}[h!]
\small
\centering
\begin{tabular}{||c|c|c|c|c|c|c||}
 \hline
 	$l_{1} \rightarrow l_{2}$ & Emb & M & F & $|Diff|$& $f_i$ & P@10 \\ [0.4ex]
 \hline\hline
 $en \rightarrow hi$ & orig  & 67.23  & 71.06  & 7.60 && 24.24\\
  & $LP_{multi}$  & 69.14 & 73.57 & 7.56 & 0.60 & 19.44 \\
  & $LP_{EQR}$  & 69.78 & 73.54 & \textbf{7.28} & 0.60 & 21.95 \\
    \hline
 $en \rightarrow be$ & orig  & 66.96 & 70.68 & 8.16 && 14.28 \\
  & $LP_{multi}$  & 69.75 & 74.37  & 7.83 & 0.35 & 18.18\\
  & $LP_{EQR}$  & 69.74 & 73.33 & \textbf{7.65}& 0.65 & 20.59\\
    \hline
 $en \rightarrow te$ & orig  & 67.30 & 70.55 & 7.58 && 21.21 \\
  & $LP_{multi}$  & 70.31  & 74.20 & 8.06 & 0.35 & 22.22\\
  & $LP_{EQR}$  & 70.59 & 74.68 & \textbf{7.30} & 0.60 & 14.29 \\
    \hline
 $hi \rightarrow en$ & orig  & 61.74 & 64.08 & 7.79 && 20.59 \\
  & $LP_{multi}$  & 62.86 & 64.83 & \textbf{7.16} & 0.47 & 16.67 \\
 & $LP_{EQR}$  & 62.31 & 64.59 & 7.22 & 0.41 & 15.63 \\
    \hline
 $hi \rightarrow be$ & orig  & 61.14 & 63.16 & 7.65 &&  \\
  & $LP_{multi}$  & 62.94 & 65.00 & 7.24 & 0.53 & - \\
  & $LP_{EQR}$  & 63.15 & 65.07& \textbf{7.19} & 0.35 & \\
    \hline
 $hi \rightarrow te$ & orig  & 61.22 & 63.49 & 7.31 && \\
  & $LP_{multi}$  & 63.19  & 65.14 & 7.13 & 0.60 &- \\
  & $LP_{EQR}$  & 61.98 & 63.86 & \textbf{6.89} & 0.53 & \\
    \hline
    
 $be \rightarrow en$ & orig  & 59.64 & 61.03 & 7.75 && 21.87\\
  & $LP_{multi}$  & 61.87 & 63.58 & \textbf{6.96} & 0.60 & 12.5 \\
  & $LP_{EQR}$  & 61.15 & 63.39 & 7.51 & 0.53 &13.89\\
    \hline
 $be \rightarrow hi$ & orig  & 58.63 & 59.63 & 7.69 & & \\
  & $LP_{multi}$  & 61.47 & 63.46 & \textbf{6.65} & 0.53 & - \\
  & $LP_{EQR}$  & 61.67 & 63.81 & 7.36 & 0.53 & \\
    \hline
 $be \rightarrow te$ & orig  & 59.42 & 61.22 & 6.91 & & \\
  & $LP_{multi}$  & 61.34 & 63.79 & 7.01 & 0.35 & -\\
  & $LP_{EQR}$ & 61.52 & 63.43 & \textbf{6.88}  & 0.53 & \\
    \hline
$te \rightarrow en$ & orig  & 57.76  & 59.77  & 7.46 & & 15.79\\
  & $LP_{multi}$  & 60.67 & 62.74 & \textbf{7.20} & 0.65 & 16.28\\
  & $LP_{EQR}$ & 61.38 & 62.46 & 8.08 & 0.47 & 33.33  \\
    \hline
 $te \rightarrow hi$ & orig  & 58.22 & 58.99 & 7.76 & & \\
  & $LP_{multi}$ & 60.48 & 62.72 & 7.52 & 0.30 &- \\
  & $LP_{EQR}$ & 61.15 & 63.10 & \textbf{6.98} & 0.35 & \\
    \hline
 $te \rightarrow be$ & orig  & 58.63  & 60.47 & 8.07 & & \\
  & $LP_{multi}$  & 61.74  & 63.42 & 7.76 & 0.53 &-\\
  & $LP_{EQR}$ & 60.96 & 62.03 & \textbf{7.72} & 0.60 & \\
   \hline
\end{tabular}
  \caption{\textit{ExBias} results in the multilingual setting. 
  LP algorithm for PCA $(k=4)$ is used for this table.}
  \label{tab:cross}
\end{table}

Table~\ref{tab:cross} summarizes the \textit{ExBias} results for the multilingual setting where we compare the multilingual extension of LP ($LP_{multi}$ and $LP_{EQR}$) with the original bilingual embedding (orig). For a bilingual embedding $l_{1} \rightarrow l_{2}$, it is evaluated on the bios of language $l_1$. Precision@10 (P@10) for cross-lingual sentence retrieval task, on \textit{Mann ki baat} (MKB) dataset is used as a multilingual-embedding quality evaluation. We observe that \textbf{(i)} $LP_{multi}$ and $LP_{EQR}$, always outperform the original embedding in terms of $|Diff|$, showing effective bias mitigation. Further, $LP_{multi}$ outperforms $LP_{EQR}$ for cases where the target language is English and a specific case of $be \rightarrow hi$. In all other settings, $LP_{EQR}$ is the winner, thus establishing the \textit{state-of-the-art performance} \textbf{(ii)} For all the settings, male, female accuracy (\textbf{M} and \textbf{F}) increases after debiasing, \textbf{(iii)} For all the language pairs, around 35-60\% professions have a reduction in performance gap ($f_{i}$), using both $LP_{multi}$ and $LP_{EQR}$ methods, \textbf{(iv)} Bias depends on the target languages, as it is better to align to gender-rich language like Telugu and Hindi than gender-neutral languages like Bengali and English and \textbf{(v)} The embedding quality (measured in terms of P@10) remains roughly same or sometimes gets a boost due to debiasing (P@10 scores only available for English\->Indic language and Indic language\->English pairs).

\if{0}

\subsection{Robustness}

\cite{gonen-goldberg-2019-lipstick} argue that projection based debiasing methods which rely on gender-direction as a notion of bias (bias-by-projection) are not able to completely remove bias. They propose that even after debiasing, the neutral words cluster together around "socially" marked male or female words (if they had male or female orientations  before debiasing). For example “nurse” being close to “receptionist”, “caregiver” and “teacher”. So percentage of male/female socially-biased words among the k nearest neighbors of the target word gives an estimate of its inclination (bias-by-neighbours). For gender-neutral profession words, they  measure the correlation between bias-by-projection and bias-by-neighbour. This correlation is expected to drop after debiasing (as socially biased words, should not have many gender-marked neighbours after debiasing). 

We replicate their proposed experiments  (i.e., Correlation between Bias-by-projection correlates to bias-by-neighbours) on embeddings for gender-neutral profession words and adjectives. We compare original embedding (before) with debiased embedding using \red{??? method (after) in Table~\ref{table:6}. We observe drop in correlation after debiasing as compared to before for all the languages considered. All these drops are significant (pvalue < 1e-4), and for Telugu, in particular the correlation has dropped to zero. This suggests are approaches are robust to ... }

\begin{table}
\scriptsize
\centering
\begin{tabular}{||c|c|c|c||}
 \hline
 Lang & $before$ & $after$ & $Drop (\%)$ \\ [0.4ex]
 \hline\hline
 $en$ & 0.541 & 0.477 & 11.78\\ 
 $hi$ & 0.440 & 0.331  & 24.71\\
 $be$ & 0.493 & 0.394 & 20.10 \\
 $te$ & 0.401 & 0.009 & 97.72\\
  \hline
\end{tabular}
\caption{Pearson correlation between bias by neighbours and bias by projection for LP-multi (PCA).}
  \label{table:6}
\end{table}

\fi
\section{Conclusion}\label{sec:conc}
In this paper, we proposed different LP variants to debias word embeddings for three Indian languages in addition to English. We experiment in two different settings - monolingual and multilingual; for each setting we measure two types of biases -- intrinsic and extrinsic. In both settings and for both types of evaluation measures we observe that debiasing helps in removing the bias from the embeddings. This comes mostly at the cost of a slight compromise of the embedding quality as is well established in the fairness literature.  

In future, we would like to extend the framework to include more Indian languages. We also plan to include other downstream tasks which at this point however is difficult due to the limited availability of datasets like BiasBios.

\bibliographystyle{ACM-Reference-Format}
\balance 
\bibliography{sample-base}


\begin{thebibliography}{21}


\ifx \showCODEN    \undefined \def \showCODEN     #1{\unskip}     \fi
\ifx \showDOI      \undefined \def \showDOI       #1{#1}\fi
\ifx \showISBNx    \undefined \def \showISBNx     #1{\unskip}     \fi
\ifx \showISBNxiii \undefined \def \showISBNxiii  #1{\unskip}     \fi
\ifx \showISSN     \undefined \def \showISSN      #1{\unskip}     \fi
\ifx \showLCCN     \undefined \def \showLCCN      #1{\unskip}     \fi
\ifx \shownote     \undefined \def \shownote      #1{#1}          \fi
\ifx \showarticletitle \undefined \def \showarticletitle #1{#1}   \fi
\ifx \showURL      \undefined \def \showURL       {\relax}        \fi
\providecommand\bibfield[2]{#2}
\providecommand\bibinfo[2]{#2}
\providecommand\natexlab[1]{#1}
\providecommand\showeprint[2][]{arXiv:#2}

\bibitem[\protect\citeauthoryear{Ahmad, Zhang, Ma, Chang, and Peng}{Ahmad
  et~al\mbox{.}}{2019}]%
        {ahmad-etal-2019-cross}
\bibfield{author}{\bibinfo{person}{Wasi~Uddin Ahmad}, \bibinfo{person}{Zhisong
  Zhang}, \bibinfo{person}{Xuezhe Ma}, \bibinfo{person}{Kai-Wei Chang}, {and}
  \bibinfo{person}{Nanyun Peng}.} \bibinfo{year}{2019}\natexlab{}.
\newblock \showarticletitle{Cross-Lingual Dependency Parsing with Unlabeled
  Auxiliary Languages}. In \bibinfo{booktitle}{\emph{Proceedings of the 23rd
  Conference on Computational Natural Language Learning (CoNLL)}}.
  \bibinfo{publisher}{Association for Computational Linguistics},
  \bibinfo{address}{Hong Kong, China}, \bibinfo{pages}{372--382}.
\newblock
\urldef\tempurl%
\url{https://doi.org/10.18653/v1/K19-1035}
\showDOI{\tempurl}


\bibitem[\protect\citeauthoryear{Ammar, Mulcaire, Tsvetkov, Lample, Dyer, and
  Smith}{Ammar et~al\mbox{.}}{2016}]%
        {Ammar2016MassivelyMW}
\bibfield{author}{\bibinfo{person}{Waleed Ammar}, \bibinfo{person}{George
  Mulcaire}, \bibinfo{person}{Yulia Tsvetkov}, \bibinfo{person}{Guillaume
  Lample}, \bibinfo{person}{Chris Dyer}, {and} \bibinfo{person}{Noah~A.
  Smith}.} \bibinfo{year}{2016}\natexlab{}.
\newblock \showarticletitle{Massively Multilingual Word Embeddings}.
\newblock \bibinfo{journal}{\emph{ArXiv}}  \bibinfo{volume}{abs/1602.01925}
  (\bibinfo{year}{2016}).
\newblock


\bibitem[\protect\citeauthoryear{Blodgett, Barocas, Daum{\'e}~III, and
  Wallach}{Blodgett et~al\mbox{.}}{2020}]%
        {blodgett-etal-2020-language}
\bibfield{author}{\bibinfo{person}{Su~Lin Blodgett}, \bibinfo{person}{Solon
  Barocas}, \bibinfo{person}{Hal Daum{\'e}~III}, {and} \bibinfo{person}{Hanna
  Wallach}.} \bibinfo{year}{2020}\natexlab{}.
\newblock \showarticletitle{Language (Technology) is Power: A Critical Survey
  of {``}Bias{''} in {NLP}}. In \bibinfo{booktitle}{\emph{Proceedings of the
  58th Annual Meeting of the Association for Computational Linguistics}}.
  \bibinfo{publisher}{Association for Computational Linguistics},
  \bibinfo{address}{Online}, \bibinfo{pages}{5454--5476}.
\newblock
\urldef\tempurl%
\url{https://doi.org/10.18653/v1/2020.acl-main.485}
\showDOI{\tempurl}


\bibitem[\protect\citeauthoryear{Bolukbasi, Chang, Zou, Saligrama, and
  Kalai}{Bolukbasi et~al\mbox{.}}{2016}]%
        {article}
\bibfield{author}{\bibinfo{person}{Tolga Bolukbasi}, \bibinfo{person}{Kai-Wei
  Chang}, \bibinfo{person}{James Zou}, \bibinfo{person}{Venkatesh Saligrama},
  {and} \bibinfo{person}{Adam Kalai}.} \bibinfo{year}{2016}\natexlab{}.
\newblock \showarticletitle{Man is to Computer Programmer as Woman is to
  Homemaker? Debiasing Word Embeddings}. In
  \bibinfo{booktitle}{\emph{Proceedings of NeurIPS}}.
\newblock


\bibitem[\protect\citeauthoryear{Caliskan, Bryson, and Narayanan}{Caliskan
  et~al\mbox{.}}{2017}]%
        {Caliskan}
\bibfield{author}{\bibinfo{person}{Aylin Caliskan}, \bibinfo{person}{Joanna
  Bryson}, {and} \bibinfo{person}{Arvind Narayanan}.}
  \bibinfo{year}{2017}\natexlab{}.
\newblock \showarticletitle{Semantics derived automatically from language
  corpora contain human-like biases}.
\newblock \bibinfo{journal}{\emph{Science}}  \bibinfo{volume}{356}
  (\bibinfo{date}{04} \bibinfo{year}{2017}), \bibinfo{pages}{183--186}.
\newblock


\bibitem[\protect\citeauthoryear{Chen, Awadallah, Hassan, Wang, and
  Cardie}{Chen et~al\mbox{.}}{2019}]%
        {chen-etal-2019-multi-source}
\bibfield{author}{\bibinfo{person}{Xilun Chen}, \bibinfo{person}{Ahmed~Hassan
  Awadallah}, \bibinfo{person}{Hany Hassan}, \bibinfo{person}{Wei Wang}, {and}
  \bibinfo{person}{Claire Cardie}.} \bibinfo{year}{2019}\natexlab{}.
\newblock \showarticletitle{Multi-Source Cross-Lingual Model Transfer: Learning
  What to Share}. In \bibinfo{booktitle}{\emph{Proceedings of the 57th Annual
  Meeting of the Association for Computational Linguistics}}.
  \bibinfo{publisher}{Association for Computational Linguistics},
  \bibinfo{address}{Florence, Italy}, \bibinfo{pages}{3098--3112}.
\newblock
\urldef\tempurl%
\url{https://doi.org/10.18653/v1/P19-1299}
\showDOI{\tempurl}


\bibitem[\protect\citeauthoryear{De-Arteaga, Romanov, Wallach, Chayes, Borgs,
  Chouldechova, Geyik, Kenthapadi, and Kalai}{De-Arteaga et~al\mbox{.}}{2019}]%
        {10.1145/3287560.3287572}
\bibfield{author}{\bibinfo{person}{Maria De-Arteaga}, \bibinfo{person}{Alexey
  Romanov}, \bibinfo{person}{Hanna Wallach}, \bibinfo{person}{Jennifer Chayes},
  \bibinfo{person}{Christian Borgs}, \bibinfo{person}{Alexandra Chouldechova},
  \bibinfo{person}{Sahin Geyik}, \bibinfo{person}{Krishnaram Kenthapadi}, {and}
  \bibinfo{person}{Adam~Tauman Kalai}.} \bibinfo{year}{2019}\natexlab{}.
\newblock \showarticletitle{Bias in Bios: A Case Study of Semantic
  Representation Bias in a High-Stakes Setting}. In
  \bibinfo{booktitle}{\emph{Proceedings of the Conference on Fairness,
  Accountability, and Transparency}} (Atlanta, GA, USA)
  \emph{(\bibinfo{series}{FAT* '19})}. \bibinfo{publisher}{Association for
  Computing Machinery}, \bibinfo{address}{New York, NY, USA},
  \bibinfo{pages}{120–128}.
\newblock
\showISBNx{9781450361255}
\urldef\tempurl%
\url{https://doi.org/10.1145/3287560.3287572}
\showDOI{\tempurl}


\bibitem[\protect\citeauthoryear{Dev, Li, Phillips, and Srikumar}{Dev
  et~al\mbox{.}}{2020}]%
        {Dev_Li_Phillips_Srikumar_2020}
\bibfield{author}{\bibinfo{person}{Sunipa Dev}, \bibinfo{person}{Tao Li},
  \bibinfo{person}{Jeff~M. Phillips}, {and} \bibinfo{person}{Vivek Srikumar}.}
  \bibinfo{year}{2020}\natexlab{}.
\newblock \showarticletitle{On Measuring and Mitigating Biased Inferences of
  Word Embeddings}.
\newblock \bibinfo{journal}{\emph{Proceedings of the AAAI Conference on
  Artificial Intelligence}} \bibinfo{volume}{34}, \bibinfo{number}{05}
  (\bibinfo{year}{2020}), \bibinfo{pages}{7659--7666}.
\newblock


\bibitem[\protect\citeauthoryear{Dev and Phillips}{Dev and Phillips}{2019}]%
        {Dev2019AttenuatingBI}
\bibfield{author}{\bibinfo{person}{Sunipa Dev} {and} \bibinfo{person}{Jeff~M.
  Phillips}.} \bibinfo{year}{2019}\natexlab{}.
\newblock \showarticletitle{Attenuating Bias in Word Vectors}. In
  \bibinfo{booktitle}{\emph{Proceedings of AISTATS}}.
\newblock


\bibitem[\protect\citeauthoryear{Gonen and Goldberg}{Gonen and
  Goldberg}{2019}]%
        {gonen-goldberg-2019-lipstick}
\bibfield{author}{\bibinfo{person}{Hila Gonen} {and} \bibinfo{person}{Yoav
  Goldberg}.} \bibinfo{year}{2019}\natexlab{}.
\newblock \showarticletitle{Lipstick on a Pig: {D}ebiasing Methods Cover up
  Systematic Gender Biases in Word Embeddings But do not Remove Them}. In
  \bibinfo{booktitle}{\emph{Proceedings of the 2019 Conference of the North
  {A}merican Chapter of the Association for Computational Linguistics: Human
  Language Technologies, Volume 1 (Long and Short Papers)}}.
  \bibinfo{publisher}{Association for Computational Linguistics},
  \bibinfo{address}{Minneapolis, Minnesota}, \bibinfo{pages}{609--614}.
\newblock
\urldef\tempurl%
\url{https://doi.org/10.18653/v1/N19-1061}
\showDOI{\tempurl}


\bibitem[\protect\citeauthoryear{Hassan and Alamgir}{Hassan and
  Alamgir}{2013}]%
        {comment2}
\bibfield{author}{\bibinfo{person}{Khandoker Hassan} {and}
  \bibinfo{person}{Mohammad Alamgir}.} \bibinfo{year}{2013}\natexlab{}.
\newblock \showarticletitle{A Comparative Study of Gender Sensitivity between
  English and Bengali}.
\newblock \bibinfo{journal}{\emph{Language in India}}  \bibinfo{volume}{13}
  (\bibinfo{date}{11} \bibinfo{year}{2013}), \bibinfo{pages}{200--208}.
\newblock


\bibitem[\protect\citeauthoryear{Libovick{\'y}, Rosa, and Fraser}{Libovick{\'y}
  et~al\mbox{.}}{2019}]%
        {Libovick2019HowLI}
\bibfield{author}{\bibinfo{person}{Jindřich Libovick{\'y}},
  \bibinfo{person}{Rudolf Rosa}, {and} \bibinfo{person}{Alexander~M. Fraser}.}
  \bibinfo{year}{2019}\natexlab{}.
\newblock \showarticletitle{How Language-Neutral is Multilingual BERT?}
\newblock \bibinfo{journal}{\emph{ArXiv}}  \bibinfo{volume}{abs/1911.03310}
  (\bibinfo{year}{2019}).
\newblock


\bibitem[\protect\citeauthoryear{Meng, Peng, and Chang}{Meng
  et~al\mbox{.}}{2019}]%
        {meng-etal-2019-target}
\bibfield{author}{\bibinfo{person}{Tao Meng}, \bibinfo{person}{Nanyun Peng},
  {and} \bibinfo{person}{Kai-Wei Chang}.} \bibinfo{year}{2019}\natexlab{}.
\newblock \showarticletitle{Target Language-Aware Constrained Inference for
  Cross-lingual Dependency Parsing}. In \bibinfo{booktitle}{\emph{Proceedings
  of the 2019 Conference on Empirical Methods in Natural Language Processing
  and the 9th International Joint Conference on Natural Language Processing
  (EMNLP-IJCNLP)}}. \bibinfo{publisher}{Association for Computational
  Linguistics}, \bibinfo{address}{Hong Kong, China},
  \bibinfo{pages}{1117--1128}.
\newblock
\urldef\tempurl%
\url{https://doi.org/10.18653/v1/D19-1103}
\showDOI{\tempurl}


\bibitem[\protect\citeauthoryear{Pande}{Pande}{2004}]%
        {anjalipande}
\bibfield{author}{\bibinfo{person}{Anjali Pande}.}
  \bibinfo{year}{2004}\natexlab{}.
\newblock \showarticletitle{Undoing Gender Stereotypes in Hindi}.
\newblock \bibinfo{journal}{\emph{Linguistik online}}  \bibinfo{volume}{21}
  (\bibinfo{date}{01} \bibinfo{year}{2004}).
\newblock


\bibitem[\protect\citeauthoryear{Prates, Avelar, and Lamb}{Prates
  et~al\mbox{.}}{2020}]%
        {comment}
\bibfield{author}{\bibinfo{person}{Marcelo O.~R. Prates},
  \bibinfo{person}{Pedro H.~C. Avelar}, {and} \bibinfo{person}{Lu{\'{\i}}s~C.
  Lamb}.} \bibinfo{year}{2020}\natexlab{}.
\newblock \showarticletitle{Assessing Gender Bias in Machine Translation - {A}
  Case Study with {G}oogle Translate}.
\newblock \bibinfo{journal}{\emph{Neural Computing Applications}}
  \bibinfo{volume}{32} (\bibinfo{year}{2020}), \bibinfo{pages}{6363–6381}.
\newblock


\bibitem[\protect\citeauthoryear{Pujari, Mittal, Padhi, Jain, Jadon, and
  Kumar}{Pujari et~al\mbox{.}}{2019}]%
        {pujari}
\bibfield{author}{\bibinfo{person}{Arun~K. Pujari}, \bibinfo{person}{Ansh
  Mittal}, \bibinfo{person}{Anshuman Padhi}, \bibinfo{person}{Anshul Jain},
  \bibinfo{person}{Mukesh Jadon}, {and} \bibinfo{person}{Vikas Kumar}.}
  \bibinfo{year}{2019}\natexlab{}.
\newblock \showarticletitle{Debiasing Gender Biased Hindi Words with
  Word-Embedding}. In \bibinfo{booktitle}{\emph{Proceedings of the 2019 2nd
  International Conference on Algorithms, Computing and Artificial
  Intelligence}} (Sanya, China) \emph{(\bibinfo{series}{ACAI 2019})}.
  \bibinfo{publisher}{Association for Computing Machinery},
  \bibinfo{pages}{450–456}.
\newblock


\bibitem[\protect\citeauthoryear{Ravfogel, Elazar, Gonen, Twiton, and
  Goldberg}{Ravfogel et~al\mbox{.}}{2020}]%
        {ravfogel-etal-2020-null}
\bibfield{author}{\bibinfo{person}{Shauli Ravfogel}, \bibinfo{person}{Yanai
  Elazar}, \bibinfo{person}{Hila Gonen}, \bibinfo{person}{Michael Twiton},
  {and} \bibinfo{person}{Yoav Goldberg}.} \bibinfo{year}{2020}\natexlab{}.
\newblock \showarticletitle{Null It Out: Guarding Protected Attributes by
  Iterative Nullspace Projection}. In \bibinfo{booktitle}{\emph{Proceedings of
  the 58th Annual Meeting of the Association for Computational Linguistics}}.
  \bibinfo{publisher}{Association for Computational Linguistics},
  \bibinfo{address}{Online}, \bibinfo{pages}{7237--7256}.
\newblock


\bibitem[\protect\citeauthoryear{Siripragada, Philip, Namboodiri, and
  Jawahar}{Siripragada et~al\mbox{.}}{2020}]%
        {siripragada-etal-2020-multilingual}
\bibfield{author}{\bibinfo{person}{Shashank Siripragada},
  \bibinfo{person}{Jerin Philip}, \bibinfo{person}{Vinay~P. Namboodiri}, {and}
  \bibinfo{person}{C~V Jawahar}.} \bibinfo{year}{2020}\natexlab{}.
\newblock \showarticletitle{A Multilingual Parallel Corpora Collection Effort
  for {I}ndian Languages}. In \bibinfo{booktitle}{\emph{Proceedings of the 12th
  Language Resources and Evaluation Conference}}. \bibinfo{publisher}{European
  Language Resources Association}, \bibinfo{address}{Marseille, France},
  \bibinfo{pages}{3743--3751}.
\newblock
\showISBNx{979-10-95546-34-4}
\urldef\tempurl%
\url{https://www.aclweb.org/anthology/2020.lrec-1.462}
\showURL{%
\tempurl}


\bibitem[\protect\citeauthoryear{Smith, Turban, Hamblin, and Hammerla}{Smith
  et~al\mbox{.}}{2017}]%
        {smith2017offline}
\bibfield{author}{\bibinfo{person}{Samuel~L. Smith}, \bibinfo{person}{David
  H.~P. Turban}, \bibinfo{person}{Steven Hamblin}, {and}
  \bibinfo{person}{Nils~Y. Hammerla}.} \bibinfo{year}{2017}\natexlab{}.
\newblock \showarticletitle{Offline bilingual word vectors, orthogonal
  transformations and the inverted softmax}.
\newblock \bibinfo{journal}{\emph{CoRR}}.
\newblock
\showeprint[arxiv]{1702.03859}
\urldef\tempurl%
\url{http://arxiv.org/abs/1702.038}
\showURL{%
\tempurl}


\bibitem[\protect\citeauthoryear{Zhao, Mukherjee, Hosseini, Chang, and
  Awadallah}{Zhao et~al\mbox{.}}{2020}]%
        {zhao2020gender}
\bibfield{author}{\bibinfo{person}{Jieyu Zhao}, \bibinfo{person}{Subhabrata
  Mukherjee}, \bibinfo{person}{Saghar Hosseini}, \bibinfo{person}{Kai-Wei
  Chang}, {and} \bibinfo{person}{Ahmed~Hassan Awadallah}.}
  \bibinfo{year}{2020}\natexlab{}.
\newblock \showarticletitle{Gender Bias in Multilingual Embeddings and
  Cross-Lingual Transfer}. In \bibinfo{booktitle}{\emph{ACL}}.
\newblock


\bibitem[\protect\citeauthoryear{Zhou, Shi, Zhao, Huang, Chen, Cotterell, and
  Chang}{Zhou et~al\mbox{.}}{2019}]%
        {Zhou2019ExaminingGB}
\bibfield{author}{\bibinfo{person}{Pei Zhou}, \bibinfo{person}{Weijia Shi},
  \bibinfo{person}{Jieyu Zhao}, \bibinfo{person}{Kuan-Hao Huang},
  \bibinfo{person}{Muhao Chen}, \bibinfo{person}{Ryan Cotterell}, {and}
  \bibinfo{person}{Kai-Wei Chang}.} \bibinfo{year}{2019}\natexlab{}.
\newblock \showarticletitle{Examining Gender Bias in Languages with Grammatical
  Gender}. In \bibinfo{booktitle}{\emph{Proceedings of EMNLP}}.
\newblock


\end{thebibliography}

\appendix

\end{document}